\documentclass[preprint, 12pt, authoryear]{elsarticle}

\usepackage{algorithm}
\usepackage{algpseudocode}

\newcommand{\algbar}{\hspace{-0.3em}\textbar\ }

\newcommand{\IfBar}[1]{%
  \State \algbar\textbf{if} #1 \textbf{then}%
}
\newcommand{\EndIfBar}{%
  \State \algbar\textbf{end if}%
}

\usepackage{amsmath}
\usepackage{amssymb}
\usepackage{booktabs}
\usepackage[inline]{enumitem}
\usepackage[T1]{fontenc}
\usepackage{graphicx}
\usepackage{multirow}
\usepackage{siunitx}
\usepackage{tabularx}
\usepackage{url}

\usepackage{hyperref}
\usepackage{cleveref}

\journal{Applied Energy}

\algblock{Input}{EndInput}
\algnotext{EndInput}
\algblock{Output}{EndOutput}
\algnotext{EndOutput}

\hypersetup{
    colorlinks,
    linkcolor={red!50!black},
    citecolor={blue!50!black},
    urlcolor={blue!80!black},
    pdfauthor={Muhammed \"{O}z}
}

\makeatletter
\providecommand{\doi}[1]{%
  \begingroup
    \let\bibinfo\@secondoftwo
    \urlstyle{rm}%
    doi:
    \href{http://dx.doi.org/#1}{%
      \discretionary{}{}{}%
      \nolinkurl{#1}%
    }%
  \endgroup
}
\makeatother

\newcommand{\Desc}[2]{\State \makebox[2em][l]{#1}#2}
\floatstyle{plaintop}
\restylefloat{table}

\begin{document}

\begin{frontmatter}

\title{Differentiable Power-Flow Optimization} 

\address[aff1]{Scientific Computing Center (SCC),\\ Karlsruhe Institute of Technology (KIT), \\76344 Eggenstein-Leopoldshafen, Germany}
\address[aff2]{Helmholtz AI, Karlsruhe}
\cortext[cor1]{Please address correspondence to Muhammed Öz or Markus Götz}

\author[aff1]{Muhammed Öz\corref{cor1}}
\ead{muhammed.oez@kit.edu}
\author[aff1]{Jasmin Hörter}
\ead{jasmin.hoerter@kit.edu}
\author[aff1]{Kaleb Phipps}
\ead{kaleb.phipps@kit.edu}
\author[aff1]{Charlotte Debus}
\ead{charlotte.debus@kit.edu}
\author[aff1]{Achim Streit}
\ead{achim.streit@kit.edu}
\author[aff1,aff2]{Markus Götz\corref{cor1}} 
\ead{markus.goetz@kit.edu}

\begin{abstract}
With the rise of renewable energy sources and their high variability in generation, the management of power grids becomes increasingly complex and computationally demanding. Conventional AC-power-flow simulations, which use the Newton-Raphson (NR) method, suffer from poor scalability, making them impractical for emerging use cases such as joint transmission–distribution modeling and global grid analysis. At the same time, purely data-driven surrogate models lack physical guarantees and may violate fundamental constraints.
In this work, we propose \emph{Differentiable Power-Flow (DPF)}, a reformulation of the AC power-flow problem as a differentiable simulation. \emph{DPF} enables end-to-end gradient propagation from the physical power mismatches to the underlying simulation parameters, thereby allowing these parameters to be identified efficiently using gradient-based optimization. We demonstrate that \emph{DPF} provides a scalable alternative to NR by leveraging GPU acceleration, sparse tensor representations, and batching capabilities available in modern machine-learning frameworks such as PyTorch. \emph{DPF} is especially suited as a tool for time-series analyses due to its efficient reuse of previous solutions, for N-1 contingency-analyses due to its ability to process cases in batches, and as a screening tool by leveraging its speed and early stopping capability. The code is available in the \href{https://github.com/Helmholtz-AI-Energy/differentiable-power-flow}{authors' code repository}.
\end{abstract}

\begin{graphicalabstract}
    \centering
    \includegraphics{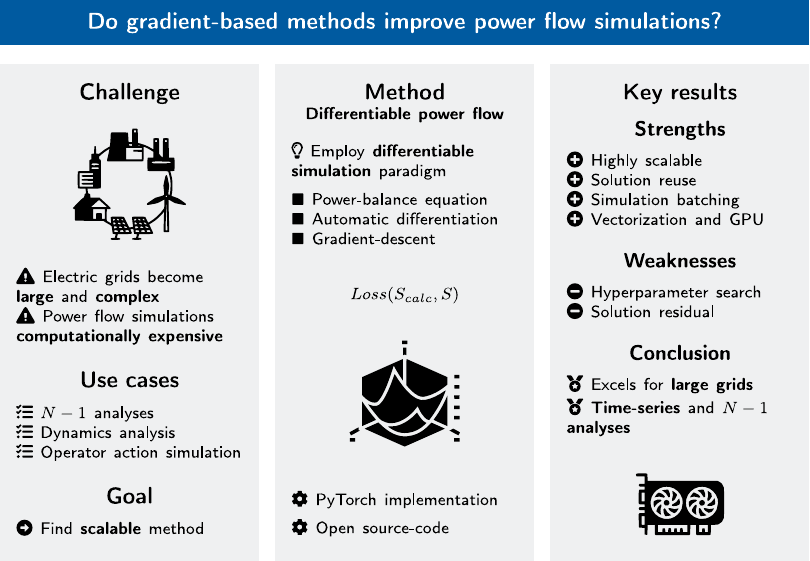}
\end{graphicalabstract}

\begin{highlights}
\item We use the concept of differentiable simulations on AC-power-flow optimizations. 
\item Our \emph{Differentiable Power-flow (DPF)} method is scalable in terms of runtime and memory usage. 
\item The calculation of multiple power-flows can be accelerated  via the GPU, making use of sparse tensors and batching.
\item When calculating power-flows for time series, \emph{DPF} is able to effectively reuse previous solutions for faster convergence.
\item \emph{DPF} can be used as a fast screening method which can be accelerated by early stopping. 
\item Easy implementation using common automatic differentiation libraries like PyTorch. 

\end{highlights}


\end{frontmatter}

\section{Introduction}
Energy grids are among the most critical infrastructures of modern society and the economy. They must be continuously monitored and actively managed to deliver electricity reliably while maintaining system stability. At the same time, power grids are inherently fragile: the failure of a single component can trigger cascading effects, as the redistribution of electrical flows may overload other parts of the network~\citep{camus2021outage, pinedo2025power, lemos2021state}. One example of an overload-induced cascade is the recent blackout that affected parts of France, Spain, and Portugal on April 28\textsuperscript{th} 2025~\citep{pinedo2025power}.

Maintaining grid stability becomes particularly challenging due to the enormous scale of modern power systems. The North American power grid alone encompasses more than 600,000 miles of transmission lines and 5.5 million miles of distribution lines~\citep{orf2023power}. Benchmark transmission models illustrate this complexity: the European transmission network represented by the \textit{case9241pegase} test case contains roughly ten thousand high-voltage buses \citep{josz2016ac}, while the North American Eastern Interconnection comprises approximately eighty thousand buses~\citep{birchfield2023review}. These figures represent only the transmission layer. When distribution networks are included, which is increasingly necessary due to the growing penetration of distributed renewable generation, the scale increases dramatically. For example, Texas alone is estimated to contain roughly 46 million electrical nodes when distribution-level infrastructure is considered as well \citep{mateo2024building}.

To ensure reliable operation of such large systems, grid operators must keep transmission line flows within thermal limits, maintain stable frequencies, and ensure that voltages remain within specified boundaries while continuously balancing electricity supply and demand~\citep{donnot2017introducing}. Achieving this becomes increasingly difficult as power systems grow in size and complexity and as uncertainty rises, for example due to weather-dependent renewable generation~\citep{hamann2024foundation}.

Simulation tools play a crucial role in the decision-making process by enabling the evaluation of different grid states in near real time. Frameworks such as \texttt{Grid2Op}~\citep{donnot2020grid2op} and
\mbox{\texttt{Pandapower}~\citep{thurner2018pandapower}} perform AC-power-flow simulations to assess system stability. These simulations enable several important applications:
For example, grid operators perform \emph{N-1 contingency analyses}~\citep{mitra2016systematic} to verify that the grid remains stable if a component such as a transmission line or substation fails. Similarly, time-series simulations are used to evaluate grid stability under changing demand and generation conditions throughout the day. To support such applications in operational settings, power-flow simulations must be computationally efficient. 

Most existing simulation frameworks rely on numerical methods such as \emph{Newton-Raphson} \emph{(NR)}~\citep{tinney1967power} to solve the non-linear AC power-flow equations. While \emph{NR} remains the state-of-the-art method due to its high accuracy, its computational cost grows rapidly with system size. As a result, \emph{NR} is not suitable for tackling emerging challenges such as the global grid~\citep{chatzivasileiadis2013global} or integrated transmission–distribution systems~\citep{Idema2013TowardsFS} and it is not suitable for expanding contingency analyses beyond N-1.

Machine-learning approaches have therefore been explored to accelerate power-flow computations. A common strategy is to train surrogate models that approximate the physical simulation directly from data~\citep{fikri2018power}. Such models can provide substantial computational speedups by eliminating the need to repeatedly solve the power-flow equations. However, because they replace the underlying physical model with a learned approximation, they may violate fundamental physical constraints and struggle to generalize reliably outside the training distribution.

To address these limitations, more recent work seeks to combine machine learning with physical laws rather than replacing the simulation entirely. One prominent approach is to incorporate physical laws directly into neural-network architectures, resulting in physics-informed neural networks~\citep{donon2020neural, bottcher2023solving, dogoulis2025kclnet}. These models improve physical consistency compared to purely data-driven surrogates. Nevertheless, they still rely on learned approximations of the system dynamics and therefore do not fully preserve the original physical consistency.

This highlights a remaining gap between traditional simulation methods and modern machine-learning approaches. What is needed is a framework that preserves the physical equations without relying on learned approximations, scales favorably to systems with millions of nodes, and preferably leverages modern hardware capabilities such as GPU acceleration, batching, and sparse operations, and integrates seamlessly with existing machine-learning infrastructure.

Rather than learning approximations of the physical system, one way to address these requirements is to make the simulation itself compatible with modern machine-learning methods~\citep{barati2025enhancing,okhuegbe2024machine,costilla2020combining}.
Differentiable simulations~\citep{liang2020differentiable, barati2025enhancing} reformulate the simulation pipeline in a fully differentiable manner, enabling gradients to be computed throughout the entire process via automatic differentiation. This preserves the underlying physical equations while enabling the use of efficient gradient-based optimization algorithms such as Adam~\citep{kingma2017adam} or stochastic gradient descent~\citep{ruder2017overviewgradientdescentoptimization}.
In the context of power networks, the power-flow problem can therefore be solved by minimizing violations of the power-balance equation derived from Kirchhoff's current law and Ohm's law (see \Cref{powerbalance}). Compared to classical second-order methods such as \emph{NR}, gradient-based optimization avoids the explicit construction and inversion of large Jacobian matrices and therefore scales more favorably with system size (see \Cref{runtime}). At the same time, differentiable simulations retain physical consistency, while naturally leveraging modern machine-learning infrastructure, including GPU acceleration, batching of multiple simulations, and sparse tensor representations.

In this work, we propose \emph{Differentiable Power-Flow (DPF)}, a reformulation of the AC power-flow problem as a differentiable simulation. This formulation enables end-to-end gradient propagation from the physical power mismatches to the underlying simulation parameters, thereby allowing these parameters to be identified efficiently using gradient-based optimization. 
We make the following contributions:

\begin{enumerate*}
    \item We introduce \emph{DPF}, applying the concept of differentiable simulations to AC power-flow calculations. Using synthetic grid data, we analyze the scaling behavior of \emph{DPF} compared to the classical \emph{NR} solver and show that \emph{DPF} achieves favorable runtime and memory scaling on large grids.
    \item We evaluate the accuracy of \emph{DPF} on standard test systems by comparing it to \emph{NR}, a highly accurate second-order solver, and to the \emph{DC power-flow approximation}~\citep{qi2012impact}, which is computationally efficient but less accurate. Our results show that the solution quality of \emph{DPF} lies between the high accuracy of \emph{NR} and the DC approximation.
    \item We demonstrate the practical applicability of \emph{DPF} for operational grid analyses, focusing on time-series simulations. To fully leverage the strengths of differentiable simulations, our implementation supports batching, GPU acceleration, warm-start initialization from previous solutions, and sparse tensor representations, which can be naturally integrated using modern machine-learning frameworks such as \emph{PyTorch}~\citep{paszke2018pytorch}.
\end{enumerate*}

\section{Background}
This section provides the foundations for our approach. First, we describe the classical power-flow equations and show how they can be expressed as an optimization problem in \Cref{sec:problem_formulation}. Next, we briefly introduce gradient-based optimization and automatic differentiation, which enable efficient gradient computation in differentiable simulations in \Cref{sec:gradient_descent}, and finally, we motivate the use of differentiable simulation techniques for power-flow calculations in \Cref{sec:differentiable_simulation}.

\subsection{Problem Formulation}
\label{sec:problem_formulation}

\begin{figure}
    \centering
    \includegraphics{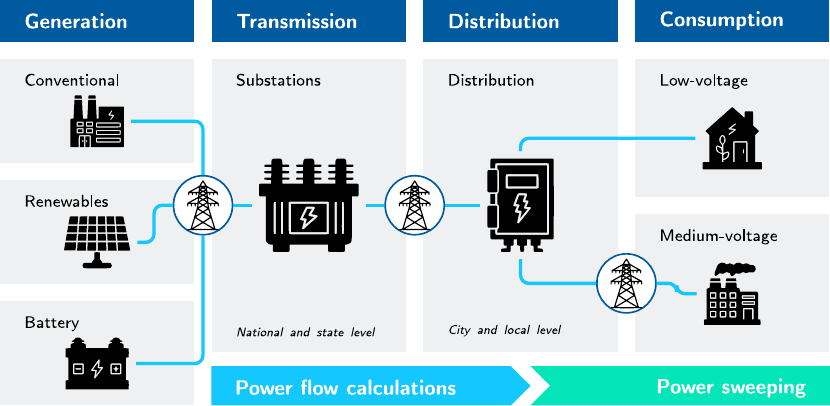}
    \caption{Schematic view of the components of the energy grid and the current applications of power flow calculations.}
    \label{fig:schema}
\end{figure}
The power grid is a complex system connecting generation and consumption through transmission and distribution layers as illustrated in \Cref{fig:schema}.
It consists of $N$ buses, or nodes, connected via transmission lines represented in the sparse admittance matrix $Y_{bus}$. $Y_{bus,ij}$ represents the admittance, i.e., the reciprocal of the impedance, between bus $i$ and bus $j$. $S_{bus,i}=P_i +iQ_i\in \mathbb{C},$ with $P_i,Q_i\in \mathbb{R,}$ denotes the complex nodal power injections at bus $i$.

Using the complete AC-power-flow model, the goal of power-flow simulations is to determine the complex voltages $V = |V|*e^{i \theta}\in \mathbb{C}^N$ such that the specified complex nodal injections $S_{bus}$ and the calculated injections $S_{calc} = V I^* = V(Y_{bus}V)^*$ match.
This leads to the non-linear power-balance equation\footnote{See \href{https://matpower.org/docs/MATPOWER-manual.pdf}{matpower-manual}, sections 3.6 and 4.1.}:
\begin{equation}
\label{powerbalance}
    S_{bus} \overset{!}{=}  S_{calc} = V I^* = V(Y_{bus}V)^*.
\end{equation}
There are three different bus types in a grid: PV-buses (generators), PQ-buses (loads) and the slack bus (generator with adaptable generation such that the power-balance equation is solvable).
Depending on the bus type, different variables among $\{|V|, \theta, P, Q\}$ are specified at bus $i$:
\begin{description}
    \item[PV-bus:] active power $P_i$ and voltage magnitude $|V_i|$
    \item[PQ-bus:] active power $P_i$ and reactive power $Q_i$
    \item[Slack bus:] voltage $V_i$
\end{description}

Computing the missing variables is the purpose of AC-power-flow solvers, while the specified variables should not change. \Cref{pf_equation_full} shows the problem without voltage boundary conditions: Find $\{V,P,Q\}$ such that 
\begin{equation}
\label{pf_equation_full}
\begin{aligned}
    P_i&={P_{calc}}_i \qquad \forall i \in pv \cup pq,\\
     Q_i&={Q_{calc}}_i \qquad \forall i \in pq,\\
     |V_i|&=|V_i|^{set} \qquad \forall i \in pv,\\
     \theta_{slack}&=0,\\
     V_{slack}&=V_{slack}^{set}.\\
\end{aligned}
\end{equation}
This can be formulated as an optimization problem:
\begin{align}\nonumber
\min_{|V|_{pq},\theta_{pv,pq}}  ||F(V)||^2 = ||S_{bus}-V(Y_{bus}V)^*||^2
\quad\textrm{ s.t. } ~ |V_i|&=|V_i|^{set} \quad \forall i \in pv,\\
\label{pf_optimization_problem}
     \theta_{slack}&=0,\\\nonumber
     V_{slack}&=V_{slack}^{set}.
\end{align}
There is a variety of different approaches to solve the optimization problem as listed in \Cref{sec:related_work}. One possible way is to use gradient-based optimization. 

\subsection{Gradient Descent Optimization}
\label{sec:gradient_descent}
Optimization problems are commonly expressed as the minimization of an objective function $f(\theta)$ with parameters $\theta\in \Theta$. The goal is to find a minimizer $\theta^*=argmin_{\theta \in \Theta }f(\theta)$.  A common approach to do this is to use gradient descent \citep{andrychowicz2016learninglearngradientdescent}. Starting from an initial parameter vector $\theta_0$, vanilla gradient descent iteratively updates the parameters by moving towards the negative gradient with a learning rate $\eta$:
\begin{equation}
    \theta_{t+1} = \theta_t - \eta \nabla f(\theta_t)
\end{equation}
Different methods exist to calculate the gradient $\nabla f(\theta_t)$, ranging from exact calculations (manual, symbolic and automatic differentiation) to numerical approximations \citep{komarov2025taxonomynumericaldifferentiationmethods,baydin2018automaticdifferentiationmachinelearning}. 
 
Symbolic differentiation works on expression trees or expression forests by applying the rules of calculus, such as the power rule, product rule, and chain rule, to obtain its derivatives \citep{zhang2025higherorderautomaticdifferentiationusing}.  It can suffer from "expression swell" generating unnecessarily large expressions \citep{zhang2025higherorderautomaticdifferentiationusing, 10.5555/60181.60188}, but this can be avoided by allowing common subexpressions \citep{laue2022equivalenceautomaticsymbolicdifferentiation}. 

A convenient and efficient way of utilizing symbolic differentiation is automatic differentiation (autodiff). Autodiff follows the idea that all numerical computations are ultimately compositions of a finite set of elementary operations for which derivatives are known. These operations form a computational graph in which derivatives can be systematically obtained using the chain rule. The essential algorithmic principle is dynamic programming: partial derivatives are stored at each node and reused, thereby avoiding redundant computations. Two main evaluation strategies exist: forward mode and reverse mode autodiff~\citep{baydin2018automaticdifferentiationmachinelearning}. In forward mode, derivatives are propagated alongside the function evaluation from inputs to outputs. In reverse mode, the function is first evaluated in a forward pass, after which partial derivatives are propagated backward from the output to the inputs. The accumulation direction is relevant. For objective functions with many parameters and a scalar output, reverse mode autodiff is particularly efficient because intermediate partial derivatives can be reused during the backward pass. This property makes reverse mode autodiff well suited for large-scale optimization problems and forms the basis of differentiable programming frameworks such as PyTorch.

As an example for reverse mode autodiff (similar to \citeauthor{Fl_gel_2025}), consider the objective function $y=f(x_1;x_2) = g_3(g_2(g_1(x_1,x_2))) \in \mathbb{R}$ with the intermediate scalar values $h_i=(g_i\circ g_{i-1}\circ \dots \circ g_{1})(x_1,x_2) \in \mathbb{R}$, $g_i:\mathbb{R} \to \mathbb{R}$ up to index $i \in \mathbb{N}$. Then the partial derivative of $y$ with regards to $x_1$ is
\begin{equation*}
    \frac{\partial y}{\partial x_1}=\frac{\partial y}{\partial h_2} \frac{\partial h_2}{\partial h_1} \frac{\partial h_1}{\partial x_1} \quad
\end{equation*}
In a computational graph representation, the variables $h_2$ and $h_1$ correspond to internal nodes, each holding their partial derivatives $\frac{\partial y}{\partial h_2}$ and  $\frac{\partial y}{\partial h_1}$. The key advantage of storing the variables is the reusability. First $\frac{\partial y}{\partial h_2}$ is calculated, then $\frac{\partial y}{\partial h_1}$ using the previously stored derivative from $h_2$. Calculating $\frac{\partial y}{\partial x_2}$ afterwards becomes trivial as $\frac{\partial y}{\partial h_1}$ is already stored inside the inner node and only one local multiplication is required. 

\subsection{Differentiable Simulations}
\label{sec:differentiable_simulation}
Physical simulations are a prime use cases for reverse mode autodiff, especially for industrial, design, engineering, and robotics applications  \citep{newbury2024review}. 
When simulations are expressed in a differentiable way, they are called differentiable simulations. Formulating a problem in this way can accelerate simulations, but also unlock new applications, e.g., learning unknown system parameters or evaluating their influence on the simulation result by solving an inverse problem \citep{fan2020solvinginverseproblemssteadystate}.  According to \citeauthor{liang2020differentiable}, a good differentiable simulator should be vectorization friendly (not contain fragmented operations during the forward pass), GPU friendly, and support sparse operations due to the locality of many physical computations. For power flow simulations, all of these criteria are satisfied with the help of automatic differentiation tools such as PyTorch.

\section{Related Work}
\label{sec:related_work}

A wide range of solutions has been proposed to perform power-flow calculations~\citep{alawneh2023review, sauter2017comparison}. Existing approaches can broadly be categorized into direct methods, iterative numerical methods, and data-driven methods.

\begin{table}
    \caption{Power-flow solution methods. Abbreviations: DC: DC-Approximation,  HELM: Holomorphic Embedding Load Flow Method, GS: Gauß-Seidel, SA: Successive Approximation, NR: Newton-Raphson, FDLF: Fast Decoupled Load Flow, DPF: Differentiable Power-Flow, GNS: Graph Neural Solver, KCLNet: Kirchhoff Current Law Network. }
    \label{tab:related-work}
    
    \small
    \begin{tabularx}{\textwidth}{lXp{2.5cm}}
        \toprule
        \textbf{Name} & \textbf{Description} & \textbf{Usage}\\
        
        \midrule
        \multicolumn{3}{c}{\textit{Direct methods}}\\
        \midrule
        
        DC approx.~\citep{qi2012impact} & Remove nonlinearities and solve linear system & Screening\\
        HELM~\citep{trias2012holomorphic} & Embed power-flow equation into holomorphic function & Guaranteed solution\\
        
        \midrule
        \multicolumn{3}{c}{\textit{Iterative methods based on fix-point-iterations}}\\
        \midrule
        
        GS~\citep{grainger1999power} & Rewrite power equation as a fix-point iteration and update buses sequentially & Educational\\
        SA~\citep{giraldo2022fixed} & Rewrite power equation as fix-point iteration and update buses simultaneously & Educational\\

        \midrule
        \multicolumn{3}{c}{\textit{Iterative methods based on linearization}}\\
        \midrule
        NR~\citep{tinney1967power} & Jacobian-inversion through matrix decompositions & Operational\\
        FDLF~\citep{stott2007fast} & Simplifies NR by removing off-diagonal blocks from Jacobian & Screening\footnotemark\\
        DPF ~\citep{barati2025enhancing} and us & Gradient-based optimization, corresponds to NR with Jacobian-transpose for squared loss function and simple gradient descent & Screening, large grids\\

        \midrule
        \multicolumn{3}{c}{\textit{Data-driven methods}}\\
        \midrule
       -~\citep{fikri2018power} & Multi-layer perceptron & Fast\\
        GNS~\citep{donon2020neural} & GNN & Fast, accurate\\
       -~\citep{bottcher2023solving} & GNN & Fast, accurate\\
        KCLNet~\citep{dogoulis2025kclnet} &  GNN w. hard constraints & Fast, physical\\
        \bottomrule
    \end{tabularx}
\end{table}

\footnotetext{{Should be faster than \emph{NR} in theory and can be used for screening and large grids but in LightSim2Grid \emph{NR} using a KLU-solver is faster due to \href{https://lightsim2grid.readthedocs.io/en/latest/benchmarks.html}{implementation reasons}.}}

\subsection{Direct Methods}

Direct methods compute the solution voltage in one iteration. The most widely used direct method is the DC-approximation~\citep{qi2012impact} which simplifies the AC power-flow equations to a linear problem by neglecting reactive power effects, voltage magnitude variations, and line resistances. Due to its computational efficiency it is commonly used for screening applications such as contingency analysis, although the resulting power-flow estimates may deviate significantly from the AC solution, with typical errors on the order of 20\%~\citep{stott2009dc}.

A more accurate novel method is the Holomorphic Embedding Load Flow (HELM) method~\citep{trias2012holomorphic}. It embeds the power equation into a holomorphic function, constructs and solves a holomorphic power series, and then evaluates the series to get the resulting voltages. While HELM is very stable and can be very accurate, it needs a significant amount of time to reach accurate results comparable to  \emph{NR}~\citep{sauter2017comparison}. As a result, HELM in its base form is too slow to be used as a screening method, for decision tools and for $N-1$ analyses. Other variants such as FFHE~\citep{chiang2017novel} in combination with \emph{NR} can achieve a better performance~\citep{huang2023review}, but it remains to be seen whether a competitive implementation of HELM can outperform \emph{NR}.

\subsection{Iterative Numerical Methods}

Iterative methods solve the power-flow equation through repeated updates until convergence in contrast to direct or data-driven methods. They can use other methods as an initialization for better stability~\citep{costilla2020combining, okhuegbe2024machine}. They can also reuse solutions of similar problems. For example, \texttt{Grid2Op} reuses previous solutions in time series analyses~\citep{donnot2017introducing}. The iterative methods are based on one of two broader concepts: fix-point iterations or linearization. 

The methods based on fix-point-iterations are Gauß-Seidel (GS)~\citep{grainger1999power} and the Successive Approximation (SA) method~\citep{giraldo2022fixed}. Both rewrite the power equation as a fix-point iteration to repeatedly update voltages. They differ in their update order. GS updates the voltages sequentially, bus by bus, and uses the updated values to update the remaining voltages, while the Successive Approximation Method updates the voltages for all buses in parallel using the old voltage values. Both methods are rather conceptional and less practically relevant as they are much slower than linearization-based methods~\citep{donnot2020lightsim2grid}, especially for larger grids~\citep{sauter2017comparison}. 

Linearization-based methods approximate the nonlinear equations locally and solve a linear system at each iteration. They include Newton-Raphson (NR), Fast Decoupled Load Flow (FDLF) and gradient-based methods. The standard operational approach is the \emph{NR method}~\citep{tinney1967power}. In each iteration of \emph{NR}, a linear system

$$\Delta V = J^{-1}(-F(V)) \Leftrightarrow J \Delta V = -F(V)$$

is solved. In practice, \emph{NR} converges to accurate solutions in a few iterations (low single digit) because of its quadratic convergence~\citep{overton2017quadratic}.
However, it can be slow when applied to large grids. To speed up \emph{NR}, it is possible to further simplify the Jacobian. The \emph{FDLF} method removes the off-diagonal blocks. The key idea is that the voltage angle is strongly coupled with the real power and the voltage magnitude is coupled with reactive power while the cross-couplings are weak. As a result, not much information is lost with the simplification. \emph{FDLF} is suitable as a fast screening method. However, we do not consider this approach in our comparison as the reported times in the \texttt{LightSim2Grid} benchmark for FDLF are slightly worse than \emph{NR}~\citep{donnot2020lightsim2grid}. 

\subsection{Data-Driven Methods}

Recent research has explored the use of machine learning to approximate power-flow solutions. Early approaches employed multi-layer perceptrons to directly map system states to voltage solutions~\citep{fikri2018power}. More recent work leverages graph neural networks (GNNs), which better capture the topology of power systems~\citep{donon2020neural, bottcher2023solving}. The main advantage of these models is their low inference time once trained.  But they usually have lower accuracy and sometimes output physically implausible predictions~\citep{dogoulis2025kclnet}. To prevent this to a certain degree, physics-informed neural networks can be employed to penalize the physical violations in the loss function (soft-constraints) and project the solutions to physically viable planes (hard-constraints)~\citep{dogoulis2025kclnet}. However, the authors do not report the additional training time for hard-constraint enforcement via projections.

\subsection{Gradient-based Approaches}
Gradient-based methods represent an alternative class of iterative solvers that have received comparatively limited attention in the power-flow literature. Unlike Newton-based methods, gradient-based optimization does not require explicit Jacobian factorization and therefore yields computationally inexpensive iterations. However, they typically exhibit linear convergence rates~\citep{garrigos2024handbook}, in contrast to the quadratic convergence of NR~\citep{overton2017quadratic}.

Previous work has explored gradient-based techniques primarily as auxiliary tools, for example to improve the robustness of Newton-based solvers or escape local minima using stochastic optimization~\citep{costilla2020combining}. 
More recently, \citet{barati2025enhancing} proposed a gradient-based formulation of the power-flow problem. 
While their work provides a detailed theoretical analysis, the reported implementations operate at significantly larger runtimes (seconds instead of milliseconds for small grids such as the case-118 grid) than optimized power-flow solvers. We believe that a fair experiment must compare the methods in their optimal state. 
In this work, we investigate gradient-based power-flow optimization within a differentiable simulation framework and evaluate its practical performance using a highly optimized Newton–Raphson implementation from \texttt{LightSim2Grid}~\citep{donnot2020lightsim2grid} as a baseline. 
Our goal is to identify application scenarios in which differentiable power flow can complement or outperform state-of-the-art solvers.

\section{The Differentiable Power-Flow Method}
\label{sec:method}

\begin{figure}
    \centering
    \includegraphics{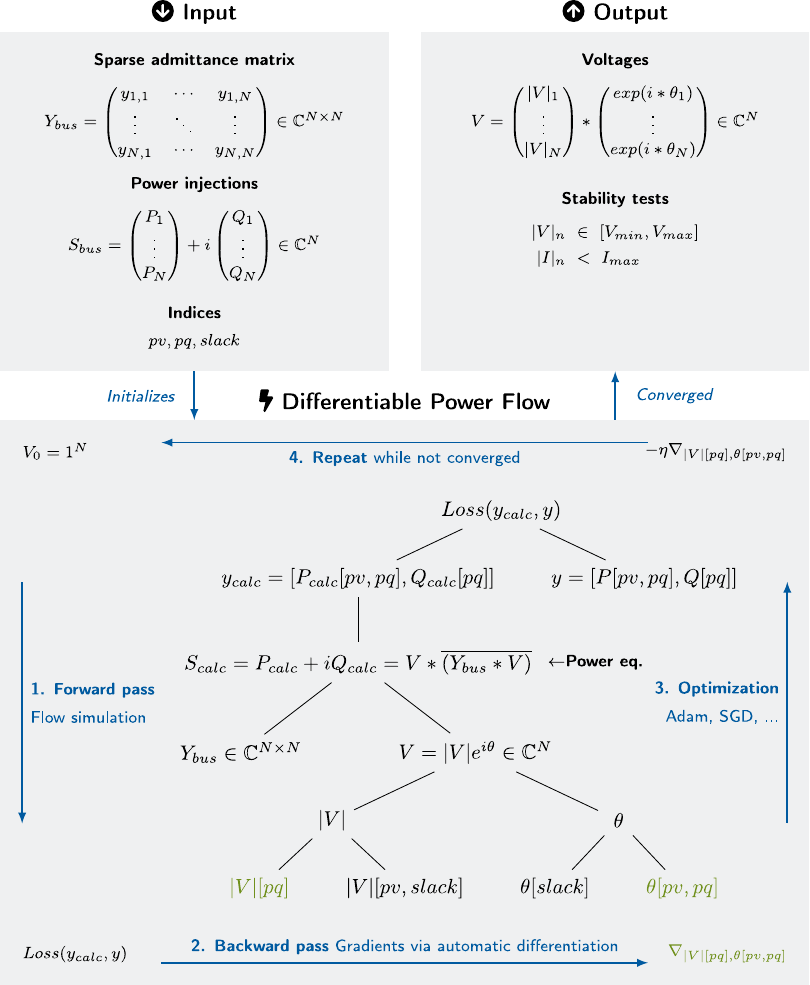}
    \caption{The figure depicts the operational graph of the power balance equation. The calculated power $S_{calc}$ (depending on the current voltage vector) and the actual power $S_{bus}$ are formed and their active/reactive components are used in $y_{calc}$ and $y$ to create the loss function. The green colored variables, namely the voltage magnitude of PV-buses and the voltage angle of PV and PQ buses, are trainable and are used as the voltage solution after training.}
    \label{fig:DPF}
\end{figure}

We propose the \href{https://github.com/Helmholtz-AI-Energy/differentiable-power-flow}{\emph{DPF}} method, which formulates the AC power-flow problem as a differentiable simulation. The key idea is to express the power-flow equations as a differentiable computational graph, enabling the use of gradient-based optimization to compute the voltage solution while preserving the underlying physical equations. 

Differentiable simulations are constructed such that every step of the numerical computation is differentiable~\citep{pargmann2024automatic,newbury2024review}. This allows gradients of the simulation output with respect to internal parameters to be computed efficiently using automatic differentiation.  Modern machine-learning frameworks such as PyTorch~\citep{paszke2018pytorch} provide built-in support for this paradigm and enable scalable implementations that leverage GPU acceleration, batching, and sparse tensor operations.

An illustration of using this approach on power-flow simulations is depicted in \Cref{fig:DPF}. Using the paradigm of differentiable simulations in power-flow studies can be done in the following way: 
The power-balance equation (\Cref{powerbalance}) defines the forward simulation. 
The complex bus voltages are represented in polar form as
\[
V = |V| e^{i\theta}.
\]
The unknown voltage magnitudes and phase angles are treated as optimization variables, while grid-specific parameters such as the sparse admittance matrix $Y_{bus}$ remain fixed.
The objective function measures the mismatch between the specified and calculated power injections,
\[
\mathcal{L}(V) = \frac{1}{N}\|S_{bus} - V(Y_{bus}V)^*\|^2 ,
\]
which corresponds to the mean squared violation of the power-balance equation. 
Minimizing this loss therefore yields voltages that satisfy the AC power-flow equations. Starting from an initial voltage estimate, the method iteratively performs forward and backward passes through the computational graph.  During the forward pass, PyTorch constructs a dynamic computational graph that records all tensor operations. By invoking backpropagation in the backward pass, PyTorch then applies reverse-mode automatic differentiation to traverse this graph and compute gradients of the loss with respect to the trainable voltage variables. These gradients are subsequently used by an optimizer to update the voltage vector.
Detailed pseudocode for our approach can be found in the appendix (\Cref{alg:dpf}).

This approach, by design, follows the characteristics of a good simulator as defined by \citet{liang2020differentiable}: it is vectorization friendly, it is GPU-friendly and it supports sparse operations.

\subsection{Implementation Details}

In our DPF implementation (see \ref{alg:dpf} in the Appendix) we make use of the classical PyTorch training loop\footnote{\url{https://docs.pytorch.org/tutorials/beginner/introyt/trainingyt.html}}. In each iteration, we conduct a forward pass, calculate the resulting loss, and finally use PyTorch's automatic differentiation functionality to calculate gradients (using the reverse mode to calculate partial derivatives in the computational graph) and update the learnable components of the complex voltage (in polar form) using an optimizer and a scheduler. It should be noted that the admittance matrix $Y_{bus}$ is stored in sparse compressed row storage format (CSR) and that all variables can be easily be placed on the GPU. In applications involving repeated simulations on the same grid (e.g., time-series analyses), grid-specific quantities such as $Y_{bus}$ and bus-type indices can be reused across iterations.

\subsection{Batched Computation}
\label{sec:batching}

An important advantage of the differentiable formulation is the ability to solve multiple power-flow problems simultaneously using batching. 
This is particularly relevant for applications such as time-series simulations or contingency analyses. 

To enable batching while preserving the original computational structure, the system matrices and vectors of individual simulations are concatenated. 
Specifically, the admittance matrices are combined into a block-diagonal matrix
\[
Y_{bus} = \text{blockdiag}(Y_{bus}^{(i)}),
\]
while voltage vectors $|V|$ and $\theta$ and the other vectors $S_{calc}$, $out$ and $target$ are stacked accordingly. 
The corresponding bus indices are shifted by $batch\_num * N$ to reflect the extended system size.
This formulation allows multiple power-flow problems to be solved in parallel using a single forward and backward pass, enabling efficient utilization of modern hardware accelerators.

\section{Theoretical Evaluation}
\label{comparison_theoretical}

In this section, we compare \emph{DPF} to \emph{NR} from a theoretical standpoint. We show how they are related in \Cref{Jacobian} and compare them with respect to time (\Cref{runtime}), memory (\Cref{memory_usage}) and stability (\Cref{stability}). 

\subsection{Relation to the Jacobian}
\label{Jacobian}

\emph{DPF} and \emph{NR} both use the Jacobian matrix in different ways. The Jacobian is defined as all first-order partial derivatives of the complex power mismatch $F(V) = S_{calc}(V)-S_{bus}$:

\begin{equation}
    J = J(V) = \frac{\partial F(V)}{\partial V} = \begin{pmatrix}
\frac{\partial P}{\partial \theta} & \frac{\partial P}{\partial \|V\|}\\
\frac{\partial Q}{\partial \theta} & \frac{\partial Q}{\partial \|V\|}
\end{pmatrix}
\end{equation}
It is a linear approximation of how the power changes with respect to changes in the voltage. 

\emph{NR} uses the Jacobian to find the voltage update that produces a power equal to the negative power mismatch. This is done by solving the linear system
\begin{equation}
    \Delta V = J^{-1}(-F(V)) \Leftrightarrow J \Delta V = -F(V)
\end{equation}
The inversion is usually done with sparse $LU$-decompositions. The new factors $L$ and $U$ can have a larger amount of non-zero elements than the Jacobian due to the fill-in. 

In contrast, gradient-based methods do not jump towards the solution immediately but update the voltage along the direction of the steepest descent. Since a scalar loss is used instead of a full mismatch vector and the corresponding Jacobian matrix, gradient-based methods use less information per iteration. For regular gradient descent with the squared error loss function 
\begin{equation}
    L(V) = \frac{1}{2} \|F(V)\|^2 = \frac{1}{2}F(V)^TF(V)
\end{equation}
the voltage update vector is
\begin{equation}
    \Delta V = - \eta \frac{\partial (L(V))}{\partial V} = -\eta J^TF(V) 
\end{equation}
This can be understood as projecting the power mismatch $F(V)$ back into voltage space through $J^T$ (instead of the inverse $J^{-1}$ in \emph{NR}). This approximation has the advantage that the Jacobian's sparsity can be used effectively without any fill-in. The downside is that the update vector is less accurate and needs to be controlled by a learning rate $\eta$.

\subsection{Run-time and Convergence}
\label{runtime}

\emph{NR} is well known to converge locally quadratically~\citep{overton2017quadratic} while gradient descent shows global linear convergence~\citep{garrigos2024handbook}. This, however, comes with a cost. \emph{NR} repeatedly solves a linear system using a large Jacobian matrix. To do this, an $LU$-decomposition can be used~\citep{donnot2020lightsim2grid}. While the worst case run time for standard $LU$-decompositions is cubic, it is possible to exploit the sparsity of the Jacobian to improve run times, e.g., using \emph{KLU}~\citep{davis2010algorithm}. The actual run time depends on the fill-in (additional \emph{number of non-zero entries (nnz)} in the decomposition compared to the sparse Jacobian). If no fill-in is present ( meaning that $nnz(J) = nnz(L) + nnz(R)$), then an iteration of \emph{NR} is in $\mathcal{O}(nnz(J))$ as only $nnz(J)$ operations are needed to create the factorization and to solve the system.

In comparison, \emph{DPF} does not have to solve a linear system. For standard gradient descent with a squared loss function the gradient can be formulated as $\frac{\partial L(V)}{\partial V} = J^TF(V)$. So the cost is dominated by a matrix vector multiplication with no fill-in with an asymptotic runtime of $\mathcal{O}(nnz(J)) = \mathcal{O}(nnz(Y_{bus}))$, which is linear for power grids.

Overall the total runtime is a tradeoff between the number of iterations needed and the time per iteration. Since this analysis does not account for different optimizers and loss functions, and since the fill-in is unclear, we evaluate the runtime experimentally in \Cref{sec:experiments}.

\subsection{Memory Usage}
\label{memory_usage}

In the following, as defined previously, let $N$ be the number of busses (nodes) and $M$ the number of connections in the grid (edges).
Aside from existing inputs (admittance matrix $Y_{bus}$, injections $S_{bus}$ and indices $pv$, $pq$ and $slack$ and the voltage vector for the solution) \emph{DPF} needs to store the complex gradient vector $\frac{\partial L(V)}{\partial V} = J^TF(V)$ consisting of $\frac{\delta LOSS}{\delta \|V\|}$ and $\frac{\delta LOSS}{\delta \theta}$, and the complex output vector created from indexing $S_{calc}$. By using reverse-mode AD it is not necessary to fully store the Jacobian as a matrix. Instead, the partial derivatives are applied directly in matrix-vector multiplication. In total, additional memory of $2N$ complex numbers is required while temporarily partial derivatives of size $M$ are used but not stored. 

As a comparison, \emph{NR} has to store the Jacobian consisting of the partial derivatives $\frac{\delta P}{\delta \|V\|}$, $\frac{\delta P}{\delta \theta}$, $\frac{\delta Q}{\delta \|V\|}$ and $\frac{\delta Q}{\delta \theta}$ ($4M$ real numbers as each partial derivative mirrors the sparsity pattern of the admittance matrix), the $LU$-factorization of the Jacobian (depends on the fill-in) and the complex mismatch vector ($N$ complex numbers). We can further simplify this by assuming $M \approx 3N$. This is the case because nodes in the grid (substations) are connected locally and have only few connections. Doing this, \emph{NR} needs to store a total of about $12N$ complex numbers as well as the fill-in from the $LU$-decomposition. 

Overall, \emph{DPF} needs about one sixth of the additional memory \emph{NR} does with no fill-in. However, with fill-in, the additional memory of \emph{NR} can be $\mathcal{O}(N^2)$.

\subsection{Stability} 
\label{stability}
\emph{NR} can be problematic when the starting point is badly chosen or when the method gets stuck in local minima. \citeauthor{costilla2020combining} solve these problems for very small grids by using stochastic gradient descent for the initialization and for cases where \emph{NR} gets stuck in local minima. They argue that, even though the Jacobian in \emph{NR} can lose its full rank, it is often still close to its full rank, which means that the gradient can still provide a useful direction~\citep{costilla2020combining}. 


\section{Experimental Evaluation}
\label{sec:experiments}

In this section, we evaluate the proposed \emph{DPF} method with a focus on its scalability and performance in large-scale settings. 
We consider two scenarios: (i) power-flow computations on increasingly large grids to analyze scaling behavior, and (ii) time series simulations, where we demonstrate how \emph{DPF} benefits from solution reuse and batching.

We explain the experimental setting in \Cref{sec:framework} and \Cref{sec:hardware} and we compare it to established methods in \Cref{sec:comparison}.

\subsection{Simulation Framework and Dataset}
\label{sec:framework}
We use the simulation framework \texttt{Grid2Op} and its backend \texttt{LightSim2Grid} \citep{donnot2020lightsim2grid} to represent power grids and conduct power-flow simulations via NR. The grid sizes range from 118 buses with the IEEE-118 grid modelling the old American power system~\citep{pena2017extended} to 9,241 buses with the case9241pegase grid modeling the European system~\citep{josz2016ac} similar to the benchmark experiments in \texttt{LightSim2Grid}\footnote{\urlstyle{same}\url{https://lightsim2grid.readthedocs.io/en/latest/benchmarks_grid_sizes.html}} (see \cref{tab:timings} for a table with the used grids). The dataset we use is the \emph{Learning Industrial Physical Simulation benchmark suite (LIPS)} ~\citep{leyli2022lips} for the 118-bus grid. It contains a data-set of different grid states suited for evaluating different power-flow techniques, especially data-driven approaches. For the other grids we use the synthetic data as used in the benchmark experiments.

\subsection{Hardware and Software Environment}
\label{sec:hardware}
Our differentiable simulation is implemented in PyTorch~\citep{paszke2018pytorch}, a popular deep learning framework, including PyTorch's sparse library to represent admittance matrices in a sparse compressed row storage (csr) format. 
The experiments were conducted on a gpu4-node with four NVIDIA A100-40 GPUs with 40GB VRAM and dual socket Intel Xeon Platinum 8,368 CPUs. We utilized CUDA 12.4~\citep{nickolls2008scalable}, NVIDIA's parallel computing platform, and PyTorch 2.6 \citep{paszke2018pytorch} to leverage GPU acceleration for faster simulations.

\subsection{Hyperparameters}
\label{sec:hyperparameter_search}
To find suitable hyper-parameters on the 118-bus-grid, we used Optuna~\citep{akiba2019optuna}. We considered the optimizers Adam~\citep{kingma2017adam}, SGD~\citep{ruder2017overviewgradientdescentoptimization} and RMSprop~\citep{hinton2012rmsprop} and the learning rate schedulers constant, step-lr, reduce-lr-on-plateau, and multi-step-lr with their respective hyper-parameters and conducted 100 trials for each combination. 
All three optimizers achieve comparable losses. So we went ahead with Adam with a reduce-lr-on-plateau scheduler. 
We noticed that the 9,241-bus-grid needs a smaller learning rate and manually lowered it. For the time series setting on the large grid, we conducted another hyperparameter search. \Cref{tab:hyperparameters-dpf} and  \Cref{tab:hyperparameters-time-series} show the hyperparameters used in a static setting and for time series. For time series, while reusing previous solutions, we noticed that the solutions are already close and a smaller learning rate is sufficient to learn the delta between time steps.

\begin{table}
    \caption{Hyperparameters for \emph{DPF} on the small 118-bus and the large 9,241-bus grid.}
    \label{tab:hyperparameters-dpf}

    \small
    \begin{tabularx}{\textwidth}{Xlcc}
        \toprule
        \multicolumn{2}{c}{} & \textbf{IEEE118} & \textbf{Case9241pegase}\\
        \midrule
        \textbf{Optimizer} & & \multicolumn{2}{c}{Adam}\\
        \midrule
        \multirow{2}{*}{\textbf{Optimizer parameters}} & lr ($\eta$) & $0.0034$ & $0.0001$\\
        & decay ($\beta$) & \multicolumn{2}{c}{$(0.979, 0.963)$}\\
        \midrule
        \textbf{Scheduler} & & \multicolumn{2}{c}{Reduce lr on plateau}\\
        \midrule
        \multirow{4}{*}{\textbf{Scheduler parameters}} & factor & \multicolumn{2}{c}{$0.547$}\\
        & patience & \multicolumn{2}{c}{$41$}\\
        & threshold & \multicolumn{2}{c}{$0.0673$}\\
        & cooldown & \multicolumn{2}{c}{$97$}\\
        \midrule
        \textbf{Maximum iterations} & & \multicolumn{2}{c}{$1,000$}\\
        \bottomrule
    \end{tabularx}
\end{table}

\begin{table}
    \caption{Hyperparameter settings for Time Series on IEEE118 with time step $t$.}
    \label{tab:hyperparameters-time-series}
    \small
    
    \begin{tabularx}{\textwidth}{Xlcc}
        \toprule
        \multicolumn{2}{c}{} & \textbf{IEEE118 $t=0$} & \textbf{IEEE118 $t>0$}\\
        \midrule
        \textbf{Optimizer} & & \multicolumn{2}{c}{Adam}\\
        \midrule
        \multirow{2}{*}{\textbf{Optimizer parameters}} & lr ($\eta$) & $0.03564$ & $0.00027$\\
        & decay ($\beta$) & $(0.9802,0.9440)$ & $(0.7847,0.6624)$\\
        \midrule
        \textbf{Scheduler} & & Step lr & Reduce lr on plateau\\
        \midrule
        \multirow{6}{*}{\textbf{Scheduler parameters}} & step & $100$ & \\
        & $\gamma$ & $0.773$ & \\
        & factor & & $0.8$\\
        & patience & & $2$\\
        & threshold & & $0.0388$\\
        & cooldown & & $4$\\
        \midrule
        \textbf{Maximum iterations} & & $1,000$ & $300$\\
        \bottomrule
    \end{tabularx}
\end{table}

\subsection{Comparison}
\label{sec:comparison}

In this section, we compare our \emph{DPF} to \emph{NR} and the \emph{DC approximation}. We test the methods for different use-cases, in particular on individual grids in \Cref{individual_pf} and on the time series application in \Cref{time_series_case}. We further compare the scalability of \emph{DPF} and \emph{NR} in \Cref{scaling_behavior} evaluating the approaches for relevant sizes needed in larger continental transmission grids and joint transmission-distribution systems.

\subsubsection{Single-Step Optimization}
\label{individual_pf}
\begin{figure}[htbp]
    \centering
    \includegraphics{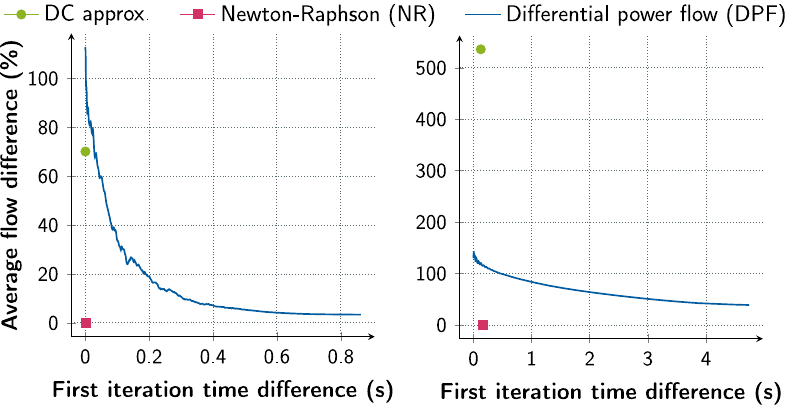}
    
    \caption{Comparison of our Differentiable Simulation (blue), NR (red) and the DC-approximation (green) on CPU without data loading time. Left: the IEEE-118 grid, right: \texttt{case9241pegase} grid. In both cases our differentiable simulation is slower than NR but the solution quality lies between NR and DC.}
    \label{fig:comparison_pareto}
\end{figure}

We begin with the standard setting of solving individual power-flows on small test grid cases. While this scenario favors second-order methods such as \emph{NR} due to their fast local convergence, it provides a useful baseline to analyze the behavior and properties of \emph{DPF}. Here, the traditional NR approach works best (see \Cref{fig:comparison_pareto}).  While our approach converges to a solution that is between NR and the DC-approximation in terms of quality, it takes about 1,000 iterations (about \SI{0.8}{\second} for the IEEE-118 and about \SI{5}{\second} for the larger 9,241-bus grid) on CPU, which is longer than the runtime of NR and DC. Despite the longer runtime for small grids, the runtime for larger grids changes in favor of \emph{DPF}.

\begin{figure}[htbp]
    \centering

    \includegraphics{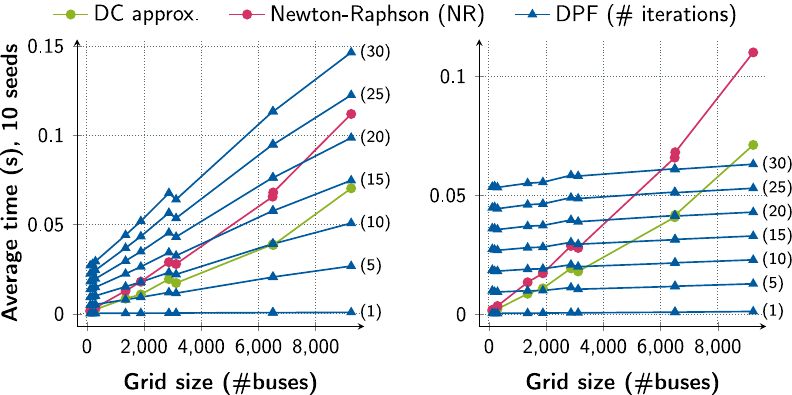}
    
    \caption{Scaling behavior of DPF (braces show the number of iterations), NR and DC. While on CPU (left) the scaling seems similar between the different approaches, on GPU (right) the better scaling behavior becomes apparent. Yet for a grid size of 9,241 NR is still faster, as our gradient-based approach needs about 1,000 iterations in its base form for convergence.}
    \label{fig:scalability}
\end{figure}

This is more apparent when looking at different grid sizes. \Cref{fig:scalability} shows the scaling behavior of our approach compared to NR and DC. The larger the grid, the more iterations of our differentiable simulation are possible. However, even for the largest available grid in the \texttt{LightSim2grid} benchmark tests (\texttt{case9241pegase}), NR is still faster. To utilize the scaling behavior more effectively, it is possible to consider multiple power-flow simulations at the same time.

\subsubsection{Time Series Optimization}
\label{time_series_case}

One use-case where \emph{DPF} shows advantages over NR is the time series setting with fixed grids and changing injections. The advantage of \emph{DPF} is the reusability of approximate solutions, e.g., power-flow solutions of previous time steps. While both \emph{NR} and \emph{DPF} are iterative methods, NR cannot fully utilize previous solutions as it still has to calculate the Jacobian matrix. On the other hand, \emph{DPF} uses many smaller steps to converge to a solution, which can be accelerated with a good initial solution. 

\begin{figure}
    \includegraphics{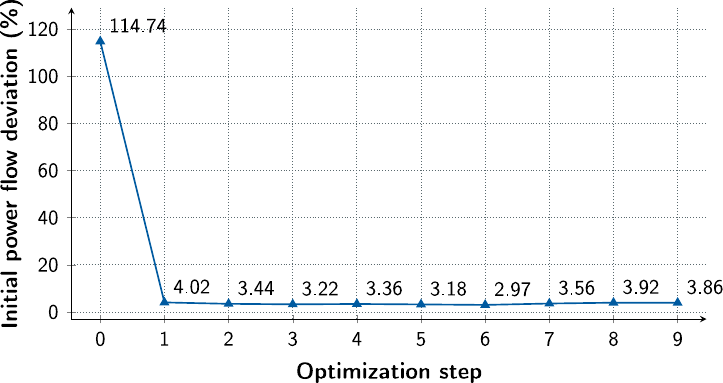}
    \caption{Solution distance of previous solutions (or initialization) to the solution of the next time step for the IEEE-118 grid. Subsequent time-steps have very similar grids, and as a result, very similar solutions.}
    \label{fig:solution distance}
\end{figure}

\begin{figure}[htbp]
    \centering
    \includegraphics{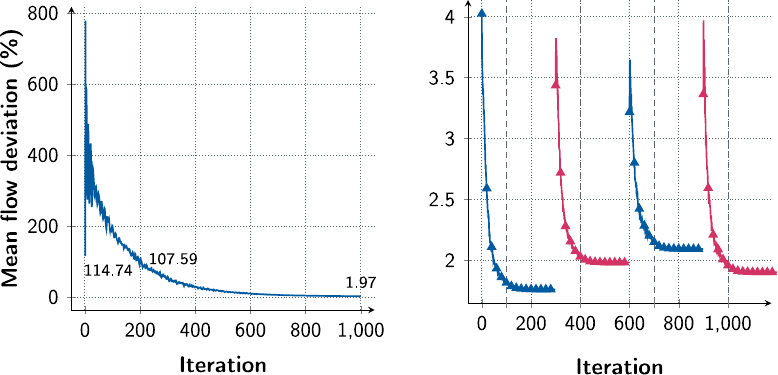}
    
    \caption{Training curves for a time series application use case for the IEEE-118 grid. The first time step (left) needs a full fitting of roughly 1,000 iterations. Albeit slow, needs to be performed only once per grid and can possibly be replaced by Newton-Raphson. Subsequent time steps (right), using the first as initial guess, only require 100 iterations to reach a plateau in solution quality.}
    \label{fig:time_series}
\end{figure}

The grid states of subsequent time steps are similar, and as a consequence, the solutions are close to each other, as shown in \Cref{fig:solution distance}. Using the similarity of solutions from subsequent time-steps we can reduce the number of iterations needed from about 1,000 to 100 iterations in \Cref{fig:time_series}. 

\begin{figure}
    \centering
    \includegraphics{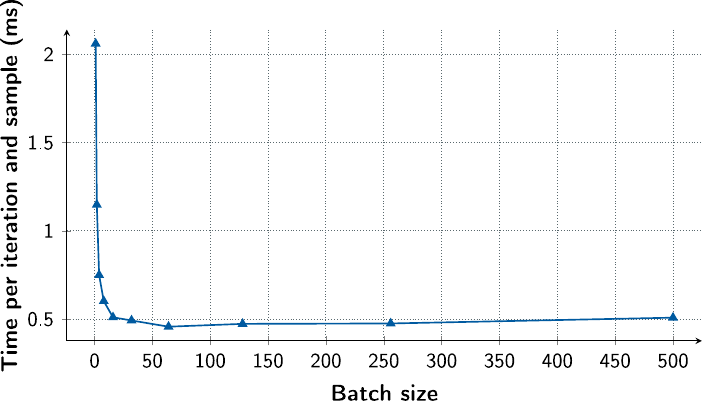}
    
    \caption{Normalized time per power-flow using batching on the \texttt{case9241pegase} grid. Without batching, a time of about \SI{2}{\milli\second} per iteration and power-flow is needed while a time of \SI{0.45}{\milli\second} per iteration and power-flow is achieved at a batch size of 64.}
    \label{fig:batching_gpu}
\end{figure}

Another advantage of \emph{DPF} that can be used in a time-series setting but is not exclusive to is the ability to use batching. 
While \emph{NR} be used to iteratively solve the subsequent power-flows one by one (and making use of information from previous time steps such as a similar admittance matrix and a similar Jacobian to reduce the calculation time), \emph{DPF} uses batching as in \Cref{sec:batching} to calculate multiple power-flows at once. Using batching on the 9,241 bus grid we improve the time per power-flow from \SI{2}{\milli\second} to \SI{0.45}{\milli\second} as shown in \Cref{fig:batching_gpu} for a batch size of 64.

\paragraph{Comparison to NR}
For the time series setting we have shown that batching and reusing previous solutions improves the simulation speed for our method. But we cannot make a fair comparison to NR as \texttt{LightSim2Grid} does not support batching. 
What we can do, however, is to compare our method to the most optimized version of NR used on time series\footnote{\urlstyle{same}\url{https://lightsim2grid.readthedocs.io/en/latest/benchmarks_grid_sizes.html}}. We ran the experiment on our device and found that the optimized version NR for time series for the \texttt{case9241pegase} grid needs \SI{12.37}{\milli\second}. This means that by the time NR is finished, our differentiable simulator has finished about 28 iterations, which is not enough to converge. Even for changing grids (and thus changing admittance matrices which DPF can handle as easily as fixed admittance matrices) NR needs \SI{26.06}{\milli\second} which is still faster than DPF (DPF can do about 58 out of 100 iterations in that time). 

\subsubsection{Scaling Behavior on Synthetic Data}
\label{scaling_behavior}
The available grids we used (see \Cref{sec:framework}) are limited to 9,241 or less busses. On these grids, \emph{NR} outperforms \emph{DPF} but the scaling appears to be in favor of our approach (see \Cref{fig:scalability}). To shed light on this, we compare both methods using synthetic data. 

\begin{figure}[htbp]
    \centering
    \includegraphics{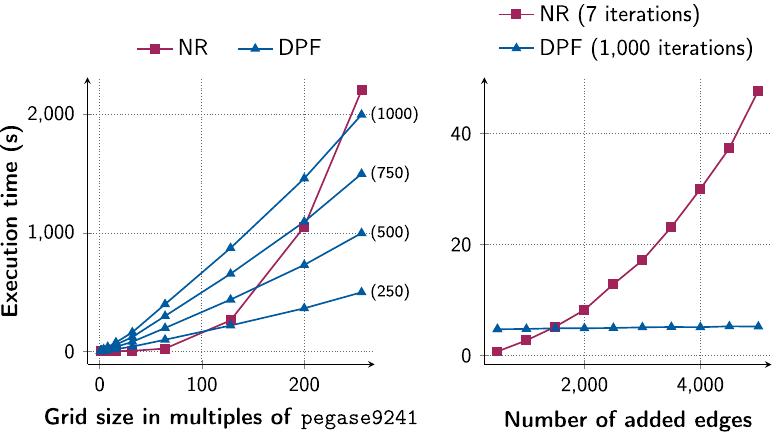}
    
    \caption{Scaling behavior with respect to grid characteristics. For node scaling (left), we increase the number of nodes in the grid by copying the \texttt{case9241pegase} grid and making few ($20* \#copies$) random connections throughout the combined grid. In case of edge scaling (right), we insert an increasing amount of random connections to the \texttt{case9241pegase} grid, adding two new non-zero entries (due to symmetry) per connection into the $Y_{bus}$ matrix consisting of $37,655$ non-zero entries. \emph{DPF} is almost unaffected (rises from \SI{4.7}{\second} to \SI{5.2}{\second}) while the runtime of \emph{NR} increases significantly.}
    \label{fig:node_and_edge_scaling}
\end{figure}

In \Cref{fig:node_and_edge_scaling} we analyze node and edge scaling. For the node scaling experiment, we increase the grid size by copying the \texttt{case9241pegase} grid and adding a few random connections to connect them. For the edge scaling experiment, we add an increasing amount of random connections to the \texttt{case9241pegase} grid. Both experiments give a similar result: \emph{NR} scaling appears to be close to quadratic while \emph{DPF} scales linearly with the number of nodes and edges. Together with the observation that the solution quality at each iteration is similar (not worse) for larger grid sizes, the node and edge scaling behavior makes \emph{DPF} a scalable approach.

\subsection{Discussion}
\label{sec:discussion}

With a differentiable simulation of the power-flow equations, gradient-based optimization techniques can be applied directly to the physical simulation. The primary advantage of \emph{DPF} lies in its scalability. 
While \emph{NR} remains highly efficient for small and medium-sized grids, the computational and memory requirements associated with constructing and solving large Jacobian matrices increase with system size. In contrast, \emph{DPF} avoids these expensive matrix factorizations and shows favorable scaling behavior in large networks. As demonstrated in our experiments, \emph{DPF} eventually surpasses \emph{NR} in convergence speed for sufficiently large systems while maintaining acceptable solution quality. 

Although classical high-voltage transmission networks alone are often too small to fully exploit this scaling advantage, the situation changes when considering broader system perspectives. The European transmission network, for example, contains roughly ten thousand buses as represented by the \texttt{case9241pegase} benchmark grid, while the North American Eastern Interconnection contains approximately eighty thousand buses~\citep{birchfield2023review}. At this scale, \emph{NR} remains highly efficient. Our results indicate that networks with on the order of millions of buses are required for \emph{DPF} to outperform \emph{NR} on the first run, although this threshold could decrease with further optimization of the method. Such system sizes become realistic when transmission and distribution networks are modeled jointly. This perspective is increasingly relevant in the context of the renewable energy transition, where distributed generation, electrification, and weather-dependent fluctuations benefit from more detailed system representations. In these settings, network sizes can reach tens of millions of nodes. For example, the electrical network in Texas has been estimated to contain approximately 46 million electrical nodes~\citep{mateo2024building}, well within the regime where scalable approaches such as \emph{DPF} become advantageous. Possible future work in this direction is to apply \emph{DPF} in a fully optimized, parallelized and perhaps distributed manner on larger realistic systems. 

Beyond its scalability, \emph{DPF} also offers additional practical benefits. Because the formulation is embedded in a differentiable programming framework, it can naturally exploit GPU acceleration and batching of multiple grid states. This makes the approach particularly suitable for applications that require solving many related power-flow problems, such as time-series simulations or $N-1$ contingency analyses. While our current implementation does not yet outperform the optimized time-series implementation of \texttt{LightSim2Grid} for small grids, the observed performance gap is relatively small, suggesting that improved implementations could become competitive.

Furthermore, gradient-based approaches can contribute to improving the robustness of traditional solvers. \emph{NR} is known to occasionally fail to converge even with good initialization~\citep{okhuegbe2024machine}. In such cases, gradient-based methods can serve as stabilizing or initialization procedures, as explored in hybrid approaches combining both techniques~\citep{costilla2020combining}.

Future work should therefore focus on optimizing the implementation of \emph{DPF}, including improved exploitation of sparsity structures, more suitable optimizers and hyperparameters, early stopping strategies, and tighter integration with existing simulation frameworks such as \texttt{LightSim2Grid}. Additionally, large-scale experiments on realistic transmission–distribution systems could further demonstrate the advantages of the approach.

\section{Conclusion}
\label{sec:conclusion}

This work demonstrates that the differentiable simulation paradigm can be successfully applied to power-flow calculations through the proposed \emph{DPF} formulation. The method preserves the underlying physical equations while enabling gradient-based optimization within modern machine-learning frameworks.

The main strength of \emph{DPF} lies in its scalability. By avoiding explicit Jacobian construction and matrix factorization, the method exhibits favorable scaling behavior as network size increases. While \emph{NR} remains the most efficient solver for small and medium-sized transmission grids, \emph{DPF} becomes increasingly attractive for very large systems where the computational and memory requirements of traditional methods become limiting. This scalability is particularly relevant for emerging applications that require large and detailed system models, such as joint transmission–distribution simulations. In such contexts, network sizes can reach millions of nodes, making scalable solution approaches essential.

Beyond scalability, \emph{DPF} offers additional practical advantages. The differentiable formulation allows seamless integration with modern machine-learning infrastructure, including GPU acceleration, batching across multiple simulations, and automatic differentiation frameworks such as PyTorch. This makes the method well suited for applications involving many related power-flow computations, such as time-series simulations and $N-1$-contingency analyses. Moreover, \emph{DPF} can provide intermediate approximate solutions that are more accurate than the widely used \emph{DC} approximation, enabling efficient screening applications where full convergence is not required.

Overall, \emph{DPF} provides a scalable alternative to classical power-flow solvers. Future work should focus on improving the efficiency and convergence behavior of the method, exploring optimized implementations, and deploying it on large realistic power-system models using parallel and distributed computing architectures.

\section*{Acknowledgements}

This work is supported by the Helmholtz AI platform grant and the Helmholtz Association Initiative and Networking Fund on the HAICORE@KIT partition. We credit Assia Benkerroum from \href{https://www.flaticon.com/authors/assia-benkerroum}{flaticon.com} for providing the icons in \Cref{fig:schema}.

\section*{Declaration of generative AI and AI-assisted technologies in the writing process.}

Statement: During the preparation of this work, the author(s) used ChatGPT to assist in the coding process and sparingly to refine the language in the document for unclear parts and explicitly not to generate new content. After using this tool/service, the author(s) reviewed and edited the content as needed and take(s) full responsibility for the content of the published article.

\clearpage

\bibliographystyle{plainnat}
\bibliography{references.bib}

\newpage
\appendix

\section*{Appendix}

\section{Baseline}
\Cref{tab:timings} shows the power-flow runtime using the NR baseline when running \href{https://github.com/Grid2op/lightsim2grid/blob/master/benchmarks/benchmark_grid_size.py}{LightSim2Grid code}. It is notable that the runtime appears to be between linear and quadratic and not cubic. 

\begin{table}
    \caption{LightSim2Grid times on our device in \si{\milli\second}. Compared are times of power-flows for changing grids with and without recycling (RC), power-flows for fixed grids on time series (TS) and times for contingency analysis (TS). The runtime scaling appears to be between linear and quadratic. The fastest times are achieved for time series as the grid (and therefore the admittance matrix and the sparsity structure of the Jacobian) stays the same and can be reused.}
    \label{tab:timings}

    \begin{tabularx}{\textwidth}{Xrrrrr}
        \toprule
        
        \multicolumn{6}{c}{\textbf{Power-flow times using NR in \si{\milli\second}}} \\
        \multicolumn{1}{c}{Grid} & \multicolumn{1}{c}{Size} & \multicolumn{1}{c}{RC} & \multicolumn{1}{c}{No RC} & \multicolumn{1}{c}{TS} & \multicolumn{1}{c}{CA} \\
        \midrule
        \texttt{case14}   & 14    & 0.0204 &   0.0491 & 0.00790 & 0.0171\\
        \texttt{case118} &   118  & 0.1180   &0.3206 & 0.0506 & 0.0746\\
        \texttt{case\_illinois200} & 200 & 0.2447 &  0.5842 & 0.1032 & 0.1705\\
        \texttt{case300}    & 300 & 0.4501 &  0.9606 & 0.2521 & 0.3341\\
        \texttt{case1354pegase} &   1,354  & 2.3303&4.0754 & 1.2324 & 1.5423\\
        \texttt{case1888rte} & 1,888  & 3.6692   &5.9012 & 1.6169&2.059\\
        \texttt{case2848rte} & 2,848  & 5.6715&9.093 & 2.447 & 3.1987\\
        \texttt{case2869pegase} & 2,869  & 5.4858 & 9.443 & 2.9107 & 3.42051\\
        \texttt{case3120sp} & 3,120  & 6.371 & 10.155 & 2.307 & 3.5356\\
        \texttt{case6495rte} & 6,495   &      17.6517    &           25.9042    &   7.53377        &       8.6814\\
        \texttt{case6515rte} & 6,515   &      20.3695    &           28.8332    &    7.3132        &       8.77377\\
        \texttt{case9241pegase} &  9,241   &      26.0554    &           41.0947    &  12.3748         &       14.0037\\
        \bottomrule
    \end{tabularx}
\end{table}

\section{Pseudocode of DPF and NR}

\Cref{alg:dpf} and \Cref{alg:newton_raphson_implementation} show pseudo-code for our method (\href{https://github.com/Helmholtz-AI-Energy/differentiable-power-flow}{\emph{DPF}}) and the baseline method 
\href{https://github.com/Grid2op/lightsim2grid}{\emph{NR}} as it is implemented.  

\begin{algorithm}

\caption{Pseudocode for our DPF method. Lines 22-26 calculate a loss from the power-balance equation which is used to update the voltages at lines 27-29 by using an optimizer and scheduler.}
\label{alg:dpf}

\small
\begin{algorithmic}[1]

\State \textbf{Hyperparameters}
\Desc{$optimizer$} \quad \quad \quad \quad optimizer with own hyperparameters
\Desc{$scheduler$} \quad \quad \quad \quad scheduler with own hyperparameters
\Desc{$loss$} \quad \quad \quad \quad loss function

\State 
\State \textbf{Inputs}
\Desc{$n$} \quad \quad \quad \quad number of active buses
\Desc{$V \in \mathbb{C}^n$} \quad \quad \quad \quad voltages $V = |V| * e^{i \theta}$ with $\theta, |V| \in \mathbb{R}^n$
\Desc{$Y_{bus} \in \mathbb{C}^{nxn}$} \quad \quad \quad \quad Admittance Matrix
\Desc{$S_{bus} \in \mathbb{C}^{n}$} \quad \quad \quad \quad Injection Vector
\Desc{pv} \quad \quad \quad \quad index list of PV-buses
\Desc{pq} \quad \quad \quad \quad index list of PQ-buses
\Desc{slack} \quad \quad \quad \quad index of slack bus
\Desc{tol} \quad \quad \quad \quad tolerance for convergence check

\State 
\State \textbf{Start}
\State $V = |V| e^{i\theta} \gets ones$ 
\Comment{Initialization}
\State $|V|_{learnable} = |V|[pq]$
\State $\theta_{learnable} = \theta[pv,pq]$
\State
\For{i=1 until max\_iter}
\State \algbar $S_{calc} = P_{calc} + i Q_{calc} \gets V(Y_{bus}V)^*$
\State \algbar $out = [P_{calc}[pv,pq], Q_{calc}[pq]]$
\Comment{Forward Pass}
\State \algbar $target = [P[pv,pq], Q[pq]]$
\State \algbar

\State \algbar $loss = MSE(out, target)$
\State \algbar $loss.backward()$
\Comment{Only update learnable parameters}
\State \algbar $optimizer.step()$
\State \algbar $scheduler.step()$

\IfBar{$loss < tol$}
\State \algbar \hspace{\algorithmicindent}\algbar \Return  $|V| e^{i\theta}$
\EndIfBar

\EndFor
\State \Return $|V| e^{i\theta}$

\end{algorithmic}
\end{algorithm}

\begin{algorithm}
\caption{NR method for power-flow calculations (LightSim2Grid~\citep{donnot2020lightsim2grid}). Every iteration a linearization is done with the Jacobian (line 22) containing the partial derivatives (lines 20-21). If the power mismatch is too large (lines 13-16, 26-29), the voltage vector is updated by the voltage delta that creates a power delta (under the Jacobian) to balance out the power mismatch (line 23).}
\label{alg:newton_raphson_implementation}

\small
\begin{algorithmic}[1]

\State \textbf{Inputs}
\Desc{$n$} \quad \quad \quad \quad number of active buses
\Desc{$V \in \mathbb{C}^n$} \quad \quad \quad \quad voltages $V = |V| * e^{i \theta}$ with angles and magnitudes $\theta, |V| \in \mathbb{R}^n$
\Desc{$Y_{bus} \in \mathbb{C}^{nxn}$} \quad \quad \quad \quad Admittance Matrix
\Desc{$S_{bus} \in \mathbb{C}^{n}$} \quad \quad \quad \quad Injection Vector
\Desc{pv} \quad \quad \quad \quad index list of PV-buses
\Desc{pq} \quad \quad \quad \quad index list of PQ-buses
\Desc{slack} \quad \quad \quad \quad index of slack bus
\Desc{tol} \quad \quad \quad \quad tolerance for convergence check

\State 
\State \textbf{Start}
\State $V = |V| e^{i\theta} \gets $ DC-power-flow 
\Comment{Initialization}
\State $S_{calc} \gets V(Y_{bus}V)^*$
\State $P_{calc}, Q_{calc} \gets Re(S_{calc}) , Im(S_{calc})$
\State $f(|V|, \theta) \gets [(P_{calc}-P)[pvpq] , (Q_{calc}-Q)[pq]]^T$
\Comment{Evaluate power mismatch}

\If{$f(|V|, \theta) < tol$}
\Comment{Update local best solution}
\State \algbar return $V$
\EndIf

\For{i=1 until max\_iter}
\State \algbar $\frac{\partial S_{calc_i}}{\partial |V|_j} \gets \left.
  \begin{cases}
    \frac{V_i Y_{bus_{ij}}^* V_j^*}{|V_j|} , & \text{for } i \neq j \\
    \frac{V_i}{|V_i|} I_i^* + \frac{V_i Y_{bus_{ij}}^* V_j^*}{|V_j|} , & \text{for } i = j \\
  \end{cases}
  \right\}  $
\Comment{Partial derivatives}

\State \algbar $\frac{\partial S_{calc_i}}{\partial \theta_j} \gets \left.
  \begin{cases}
    -iV_i Y_{bus_{ij}}^* V_j^* , & \text{for } i \neq j \\
    i V_i I_i^*  - iV_i Y_{bus_{ij}}^* V_j^* , & \text{for } i = j \\
  \end{cases}
  \right\}  $

\State \algbar $J_f \gets \begin{pmatrix}
\frac{\partial P_{calc}}{\partial \theta}[pvpq,pvpq] & \frac{\partial P_{calc}}{|V|}[pvpq,pq]\\
\frac{\partial Q_{calc}}{\partial \theta}[pq,pvpq] & \frac{\partial Q_{calc}}{|V|}[pq,pq]
\end{pmatrix}$
\Comment{Determine Jacobian}

\State \algbar $\begin{pmatrix}
\theta[pvpq]\\
|V|[pq]
\end{pmatrix} \gets \begin{pmatrix}
\theta[pvpq]\\
|V|[pq]
\end{pmatrix} - J_f^{-1}f(|V|, \theta)$
\Comment{Newton-step using linear solver}

\State \algbar $S_{calc} \gets V(Y_{bus}V)^*$
\State \algbar $P_{calc}, Q_{calc} \gets Re(S_{calc}) , Im(S_{calc})$
\State \algbar $f(|V|, \theta) \gets [(P_{calc}-P)[pvpq] , (Q_{calc}-Q)[pq]]^T$
\Comment{Power mismatch}
\IfBar{$f(|V|, \theta) < tol$}
\Comment{Update local best solution}
\State \algbar \hspace{\algorithmicindent}\algbar return $V = |V|e^{i\theta}$
\EndIfBar
\EndFor
\State return $V = |V|e^{i\theta}$

\end{algorithmic}
\end{algorithm}

\end{document}